\documentclass[letterpaper, 10 pt, conference]{ieeeconf}  

\IEEEoverridecommandlockouts                              
\usepackage{graphics} 
\usepackage{epsfig} 
\usepackage{amsmath} 
\usepackage{verbatim}
\usepackage{url} 
\title{\LARGE \bf
Sentiment Predictability for Stocks 
}

\author{Jordan Prosky$^{1}$, Andrew Tan$^{2}$,  Xingyou Song$^{3}$, Michael Zhao$^{4}$,
\thanks{*CS 294-136 Final Project, authors in alphabetical order}
\thanks{$^{1}$ jorpro@berkeley.edu}
\thanks{$^{2}$ andrewtan@berkeley.edu}
\thanks{$^{3}$ xsong@berkeley.edu}
\thanks{$^{4}$ m.zhao@berkeley.edu}
}

\begin{document}

\maketitle
\thispagestyle{empty}
\pagestyle{empty}

\begin{abstract}
In this work, we present our findings and experiments for stock-market prediction using various textual sentiment analysis tools, such as mood analysis and event extraction, as well as prediction models, such as LSTMs and specific convolutional architectures. 

\end{abstract}

\tableofcontents
\section{Introduction}

Stock market prediction is an important time-series learning problem in finance and economics. According to the semi-strong form of the efficient market hypothesis (EMH), markets are efficient and stock prices already reflect all publicly available information such as financial news and price data. This hypothesis incorporates the weak-form EMH, which implies that stock prices reflect all market information, a subset of public data. Semi-strong form EMH assumes that the market adjusts quickly to absorb new information and stock prices reflect all new available information. Given this assumption and the fact that investors purchase stocks after public information is released, an investor theoretically cannot benefit or outperform the market by trading on new information. Several studies contradict this hypothesis, and through our research, we show that we are able to outperform the market using a combination of deep learning, temporal data, and news articles.

Our report focuses on the use of various time-series and NLP techniques to extract information about an individual stock's chance of closing higher or lower than its opening price. We take advantage of the temporal nature of both stock price data and financial news data. In Section 2, we explore the current state-of-the-art in stock movement classifcation and the use of deep learning in financial prediction problems. Current advancements in sentiment analysis for stock markets focus on prediction for entire indexes. Section 3 outlines the problem of individual stock prediction and potential solutions to addressing these differences.

Our experiments, based on the state-of-the art models however show that perhaps more investigation (in network architecture, size of data, hyper-parameter tunings, and setup) is needed to produce better accuracies, and is an addition to the current literature on this topic. 

\section{Relevant and Previous Works}
Previous works in this area are sparse and may be dubious and untrustworthy. Furthermore, very few of the works specifically focus on individual companies and relative stocks. Nonetheless, we reviewed previous works. The first seminal paper relating the role of negative words in financial news documents to stock markets was \cite{words}, where the author explored the relationship between a "pessimism" index in papers to the Dow Jones Industrial Average. The most basic form of the "pessimism" was the number of negative words in the Wall Street Journal. The second work attempts to explain the relationship between negative words and individual stock pricing, found in \cite{media}. Further attempts under this work using more recent Natural Language Processing Developments were found using data such as Twitter sentiment.

\subsection{Aggregate Twitter Prediction of Dow Jones}
There has been previous work on the macro-level sentiment of Twitter posts and its relationship with the index of aggregate markets, such as the Dow Jones Average.  ~\cite{twitter}, cited 2800 times, found that there was statistical significance between the Granger Causality of certain aggregate "mood" scores of 340,000 daily Twitter posts and the Dow Jones index approximately 3-4 days later. Using a neural network, they predicted the direction ("up" or "down) with 86\% prediction rate. 

Following this work, there were numerous works studying which individual stocks from the S+P 500 indices and the Dow Jones Averages, followed the Twitter sentiment prediction more closely. These include ~\cite{twitterPLOS}, ~\cite{sentiment}  However, this is perhaps highly similar to finding which stocks have high correlation to those two average indices. Furthermore, many of the Twitter posts were irrelevant to the stocks themselves. 

Because of a still dearth of work on predicting individual stock return changes, this project was focused on experimenting with popular individual stock prediction. We explore individual stock prediction, and quantify some of these results with experiments.

\subsection{Stock Movement Prediction Using News Article Titles with SVMs}

There has been previous work on using news article titles and deep recurrent neural networks for predicting directional stock movement. Many studies contradict the efficient market hypothesis, including \cite{textual-analysis}, which achieved 57.1\% accuracy in predicting the directional movement of the S\&P 500 Index using Support Vector Machines (SVMs) trained on a feature vector containing both financial article data and stock price data over the same time period. Radial basis function (RBF) kernel SVMs are well-suited for classification problems involving numerical representations of text data and have performed well on stock market prediction based on news headlines. However, SVMs are unable to take advantage of the temporal nature of news data, which offers an avenue for improvement.

\subsection{Deep Learning in Finance}
Financial markets produce massive amounts of publicly available data every second. Financial prediction problems – such as pricing securities, constructing portfolios, and managing risk – often involve intricate data interactions that are difficult to interpret or speculate in a purely economic model. Applying deep learning models to these problems can produce more robust and useful results than traditional methods. In particular, deep learning can detect and exploit patterns in the data that are invisible to existing financial economic theory. The deep learning hierarchy has a number of advantages in prediction and classification: overfitting is more avoidable, non-linearities and complex data patterns can be accounted for, and input data can be expanded to include all possibly relevant features. In \cite{deep-learning-finance}, Heaton et al. introduced a variety of deep learning methods with applications in finance including dropout for model selection, autoencoders, and LSTM models. One application of deep learning they demonstrated was using an autoencoder to replicate and reconstruct a stock index with a subset of stocks. The sequential nature of financial data fits well with RNN models like LSTMs and GRUs. We want to extend upon existing research in deep time series analysis to create strategies for financial prediction.

\section{Individual Stock Prediction}
\subsection{Difficulties} 
While the papers may state that they used many data sources, finding these sources is not an easy task. Also, prediction using relevant individual news articles to each stock is significantly more difficult, both practically and conceptually, than aggregate Twitter $\rightarrow$ aggregate stock prices.

The practical reasons include: 
\begin{itemize} 
\item It is difficult to have a reliable method of news article scraping. Data collection is both expensive monetarily, and in many cases, websites may block any basic web-crawling. 
\item Because news article collection is difficult, and news articles may be sparse, it is therefore difficult to get a sufficient number of articles per day.  
\item In terms of reading any HTML files collected, much of it may be unreliable due to the special encodings for many website. 
\end{itemize}
The conceptual reasons also include: 

\begin{itemize}
\item News articles do not typically follow normal sentiment patterns. Many articles are written objectively and without emotion, which makes judgement by an article's "happiness, sadness", etc. unreliable. Also, a specific news site may generate an underlying bias to the sentiment. 
\item Following this, most sentiment analysis does not keep track of a "importance" score. For example, one day's volume (i.e. total trades) may be based on a significant news event, but sentiment can only process mood. It is generally known that election days for the US significantly affect stock prices, but news reports may under-report this under a sentiment score. 
\item The sparsity of articles therefore makes statistical significance problematic - there may only be sparsely 1 or 2 articles per day, compared to the 340,000 Twitter posts that can be used per day.  
\item It is not known what is the correct pattern to focus on, and the usual parameters (time lag, meta-data of stocks) are unknown. For example, one might approach this with: 1. Pos/Neg Sentiment vs. Daily (closing - opening) price, with days of lag, or 2. Intensity of article vs stock volume, or even perhaps 3. Immediate stock pricing, 1 hour after an release of an article. 
\end{itemize}

\subsection{Advanced Textual Information Approaches} \label{adv}
There are few works that attack the single-stock problem, but require significantly more sophisticated methods. \cite{DLstock} notes that news titles are actually more useful for prediction than the news content themselves, with the reference \cite{news_short} showing these experimental results. The two references both use a similar method to encode events. 
\newline
\newline 
They attack the problem of learning from a title by representing each event as $(O_{1}, O_{2}, P, T)$ tuple \label{event}, where $O_{1},O_{2}$ are two sets of objects, $P$ is a relation of the objects where $O_{1}$ "acts" $P$ on $O_{2}$, and $T$ is a time interval. An example given is: “Sep 3, 2013 - Microsoft agrees to buy Nokia’s mobile phone business for 7.2 billion.” being modeled as: (Actor = Microsoft, Action = buy, Object = Nokia’s mobile phone business, Time = Sept 3, 2013). They use OpenIE to extract all possible event tuples from a title, and 

They also noted many of the practical problems above, and addressed these issues:
\begin{itemize}
\item The most available textual information was from Bloomberg and Reuters. Furthermore, there was no gain in performance from using an entire article compared to using an article's summary according to \cite{news_short}.
\item Furthermore, the problem of sparse news was solved by encoding each summary into the event tuple ~\ref{event}, and using StanfordNLP packages to convert an event to a "generalized event", where words such as verbs were classified into a category (e.g. "bought", "purchase", "auction" $\rightarrow$ "buy").
\item It seemed that testing 1 day, 3 days, and 1-week lags were the most promising time delays from news article to stock price. 
\item Lastly, the problem of articles possessing more "dimensionality" in their meanings other than mood was solved by using event extraction for classification.
\end{itemize}

\section{Experiments}

\subsection{Stock Price and Financial News Data}

We collected and preprocessed (normalized data through the min-max method) daily stock data using Google Finance for the 30 DJIA stocks from the last 15 years. We also collected data on the S\&P 500 and Dow Jones Industrial Average indexes, which track entire industries. The daily features encompassed by our model include:
\begin{itemize}
\item Opening price: stock price at 9:30 AM ET
\item Closing price: stock price at 4:00 PM ET
\item High: highest price of the day
\item Low: lowest price of the day
\item Volume: number of shares traded during that day
\end{itemize}
For binary classification tasks, we generate labels that indicate whether a stock moved up or down intraday. We compare the opening and closing prices of the day: if the closing price was greater than or equal to the opening price, we set a label of 1, otherwise, it was given a label of 0.
For our LSTM model using stock price data, our input data is a $s$ x 5 matrix, where $s$ represents our LSTM window size and 5 is the number of daily features we collected.

\subsubsection{Websites Used}
We attempted to use the following websites: \url{nytimes.com}, \url{cnn.com}, \url{reuters.com}, \url{wsj.com}. However, only \url{reuters.com} was accessible for large-scale scraping; all other websites either blocked the scraping request, or did not properly organize news into corresponding stocks.
\newline 
Thus, we used only \url{reuters.com} news articles. 
\subsubsection{Sentiment Analyzer}
We attempted to use the following sentiment analysis tools: 
\begin{itemize}
\item VADER 
\item Google Sentiment API.
\end{itemize}

When using VADER \cite{VADER}, which is part of the NLTK for Python based on a work for internet sentiment trained on news corpus, this sentiment analysis tool has 4 scores: compound, neutral, positive, and negative. The last 3 scores sum to 1, while the compound score is a measure of the "intensity" of the input. Google Sentiment API \footnote{\url{https://cloud.google.com/natural-language/docs/analyzing-sentiment }} was also considered, however empirical tests showed that it displayed similar values with VADER, as well as being unavailable for free-to-use. 
\subsubsection{OpenIE Extractor and Word2Vec}
StanfordNLP's OpenIE Extractor can be found in \url{https://stanfordnlp.github.io/CoreNLP/openie.html}. Once it converts each sentence into a event tuple,we then used the Word2Vec Skip-Gram Algorithm, found in \url{https://pypi.python.org/pypi/word2vec}.
\subsubsection{Deep Learning Packages}
TensorFlow was used to output the event embeddings, while Keras was used for LSTM-modelling. Lastly, PyTorch was used for the convolutional network for up-down prediction.

\subsection{Setup}

\subsubsection{LSTM-Based Mood Prediction}

We use NLTK sentiment scores as well as previous prices in order to create multi-step predictions. Our method is able to work for many different companies, but for this example, we focus on 600 days of Amazon (AMZN) returns. The forecasting LSTM we use can take in an arbitrary length input sequence and predict the closing prices for the next k days. We use a simple LSTM due to the small size of our data set, with only 8 hidden units, trained for 5 epochs. We dedicate the last 50 days to be our test set to evaluate the model trained on the previous time steps. 

Although not presented, during many experiments we found that the model's predictions rely heavily on the length of the training set and the input sequence. The optimal length of training data varies over time and security, so we experimented with trial and error for the above results, where 600 days of data seemed to yield reasonable predictions. 
\newline

\subsubsection{Event Extraction using Neural Networks}
In this section, we try to represent our article titles as a single vector that we can flexibly pass into any classifier or regressor. We start by learning 100-length word embedding vectors using the Word2Vec skip-gram algorithm for the words in our news titles. With the word embeddings, we could simply average the word embeddings of all words in a news title, but we decided to use a more complex approach that should produce vectors that better capture the relationships between words in the titles. 

From StanfordNLP, we used OpenIE extraction, as seen from \cite{DLstock} and explained in \ref{adv}, to extract a sentence's underlying event structure, which for every news title yields between 0 and 10 possible event tuples. We then converted our event tuples into word embedding tuples. To get a single word embedding tuple $E$ for each news title, we averaged the up to 10 possible word embedding tuples. At this point, we fed our word embedding tuples into a neural tensor network (below in figure \ref{event_embedding}) in order extract each brief summary into a single 100-length vector event embedding.
\begin{figure}[h]
\centering

\includegraphics[scale=0.40]{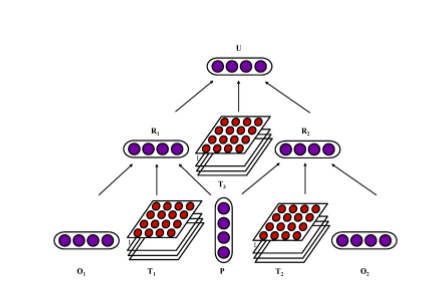}
\caption{Event embedding architecture found in \cite{DLstock}. }
\label{event_embedding}
\end{figure}

This network attempts to learn the relationships between the actor $O_1$ and action $P$ and between the action $P$ and object $O_2$ to learn $R1$ and $R2$, respectively, then learn the relationship between $R1$ and $R2$ to produce the final predicted event embedding $U$ which should capture the overall meaning of the article title. For every article, there is a learned event embedding $M$, which is a trainable parameter. To score an embedding, we take the inner product between $M$ and the predicted event embedding $U$.

As in \cite{DLstock}, we create corrupted event tuples $E^r$ where we replace the actor embedding vector with the word embedding of a random possible word, which we also pass through our network to produce corrupted event embeddings $U^r$. We can then compute the margin loss between the actual event tuple and the corrupted one with $\ell_{2}$ regularization to train our network. 
$$MarginLoss(E, E^r) = max(0, 1-M\dot{U}+M\dot{U^r})$$
$$\ell_{2} \text{-loss} = ||(T_1, T_2, T_3, W, b)||_2^2$$

We set our $\ell_{2}$ regularization weight $\lambda$ to 0.0001 and trained for 500 epochs to produce the best event embeddings as seen below in figure 2.

\begin{figure}[h]
\centering

\includegraphics[scale=0.40]{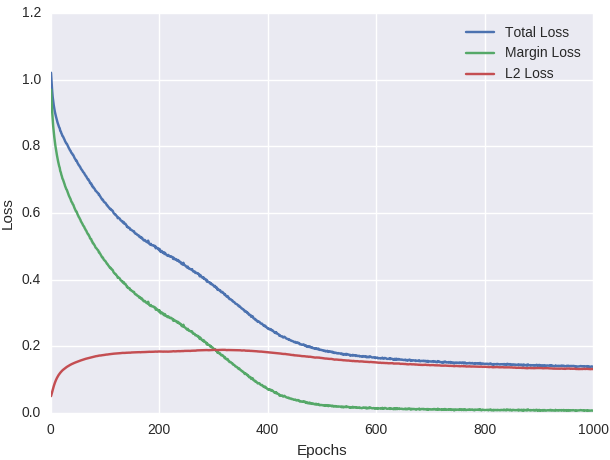}
\caption{Training losses for neural tensor network learning event embeddings. }
\label{ntn_losses}
\end{figure}

Such event embeddings may be used in traditional classifiers: SVM, Naive Bayes, etc. as well.

\subsubsection{LSTM using Event Embeddings}

Here, we have sentence embeddings for news headlines, and we use them as features with the goal of predicting whether a given company's stock will go up or down the following day. 

The data used consists of 8,000 sentence embeddings for different news article headlines, labeled with the future up or down movement of the company's stock. We use a simple LSTM model for binary classification, and see better-than-random results! We manage to consistently achieve 52\% out-of-sample accuracy, while the training accuracy continues to increase as the model is trained. This leads us to believe that we simply do not have enough data to create a robust model, as evident by the drastic over-fitting. Nonetheless, we see that there is some promise in using LSTMs as a mapping between sentence encodings and stock price movements.

\subsubsection{Convolutional Networks for Long, Medium, Short Term using Event Embeddings} \label{conv}
(Found in NN\_event\_to\_prediction.ipynb). We replicated the neural network experiments found in \cite{DLstock}, in which the word embeddings $(U_{day}, \{U_{week}^{i}, i =1,2... \}, \{U_{month}^{j}, j =1,2,... \}   )$ were injected into a feed-forward network consisting of convolutional filters for long-term $ \{U_{month}^{j}, j =1,2,... \} \rightarrow V^{month}$ (set of all events $j$ in the month before day $d$), middle-term events $\{U_{week}^{i}, i =1,2,... \} \rightarrow V^{week}$ (set of all events $i$ in the week before day $d$), concatenated with a short term (day $d$'s article $U_{day} = V^{day}$). The concatenation $V = \left[V^{month}, V^{week}. V^{day} \right]$ was inputted into a fully-connected hidden layer, with sigmoid transforms for both concatenation $\rightarrow$ hidden layer ($H = \sigma(W_{1}^{T} \cdot V )$), and hidden layer $\rightarrow$ output ($O = \sigma(W_{2}^{T} \cdot H)$). Because of the variability of the lengths of the monthly and weekly data, 0-padding was used. A soft-max was also applied on the output for the final 2-length output ($y_{class}, class \in \{up, down\}$ where $y_{class} = Softmax(O)$), in order to interpret as probabilities. This was then used to predict the up-down classification for the price change between day $d$ and day $d+1$. 

 \begin{figure}[h]
      \centering
     
      \includegraphics[scale=0.27]{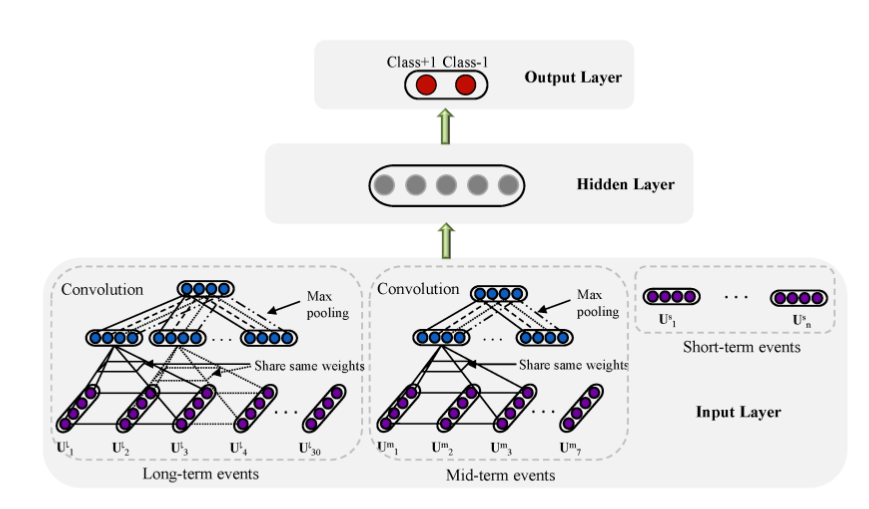}
      \caption{Feed-forward architecture found in \cite{DLstock}. Note that the middle-term and long-term events will be inputted into convolutions and max-pools to extract the best features. The network was performed in PyTorch.  }
      \label{event_NN}
   \end{figure}

We also based a network purely on the short-term events, (without convolution on past events), as a reference. 
\newline 
Because of the need for sequential processing (i.e. we should not train on a random sample and test on samples whose timing is somewhere between the random samples), we used window-ed cross validation, where $r$ of the first events were used for training, and the rest of the most recent events for testing, where $0 < r <1$ is a ratio.

\newpage 
\section{Results} 
\subsection{Data Numbers}
We present basic figures for our data. IBM and AMD were not used in the final experiments due to lack of clean data. 
 \begin{figure}[h]
      \centering
     
      \includegraphics[scale=0.35]{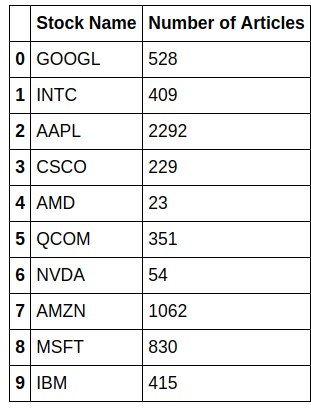}
      \caption{Number of Articles Used. }
      \label{num_articles}
   \end{figure}
\begin{figure}[h]
      \centering
     
      \includegraphics[scale=0.35]{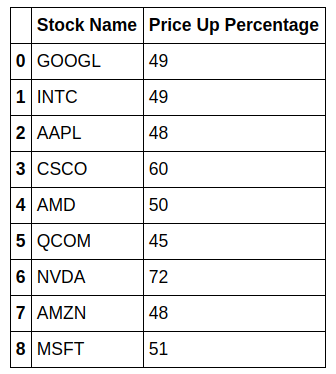}
      \caption{Percentage of news articles with stock price heading up the next day. (IBM inconclusive) }
      \label{num_articles}
   \end{figure}

\newpage 
\subsection{Basic Exploratory Analysis}
\begin{figure}[h]
      \centering
     
      \includegraphics[scale=0.35]{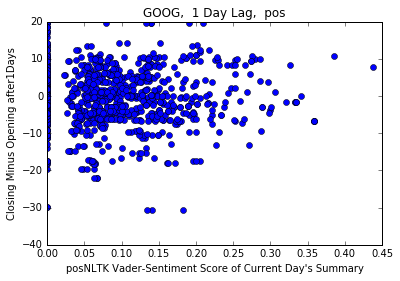}
      \caption{Price Changes Compared with NLTK Positive Scores for GOOG. }
      \label{nltk_goog}
      
      \includegraphics[scale=0.35]{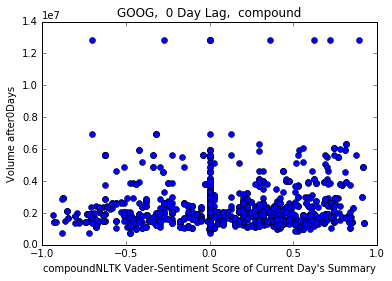}
      \caption{Stock Volumes Compared with NLTK Compound (intensity) Scores for GOOG. }
      \label{nltk_goog_comp}
   \end{figure}

The correlation between the Vader NLK sentiment scores on day $d$ of an article and the changes in price from day $d$ to the price on days $d+1, d+3, d+7$ were all inconclusive, as somewhat expected. Regardless of which score (pos, neg, neutral, or intensity), there was less than an absolute value of $0.6$ correlation between all stocks and their scores, shown in \ref{nltk_goog_comp}. 

\subsection{LSTM-Based Sentiment Score}

Below, we present the full time series of the Amazon closing prices, as well as our predictions for the last 50 days (5-day future prediction at each of the last 50 days), denoted by the red lines. 

\begin{center}
\begin{figure}[h]
\includegraphics[width = 0.5\textwidth]{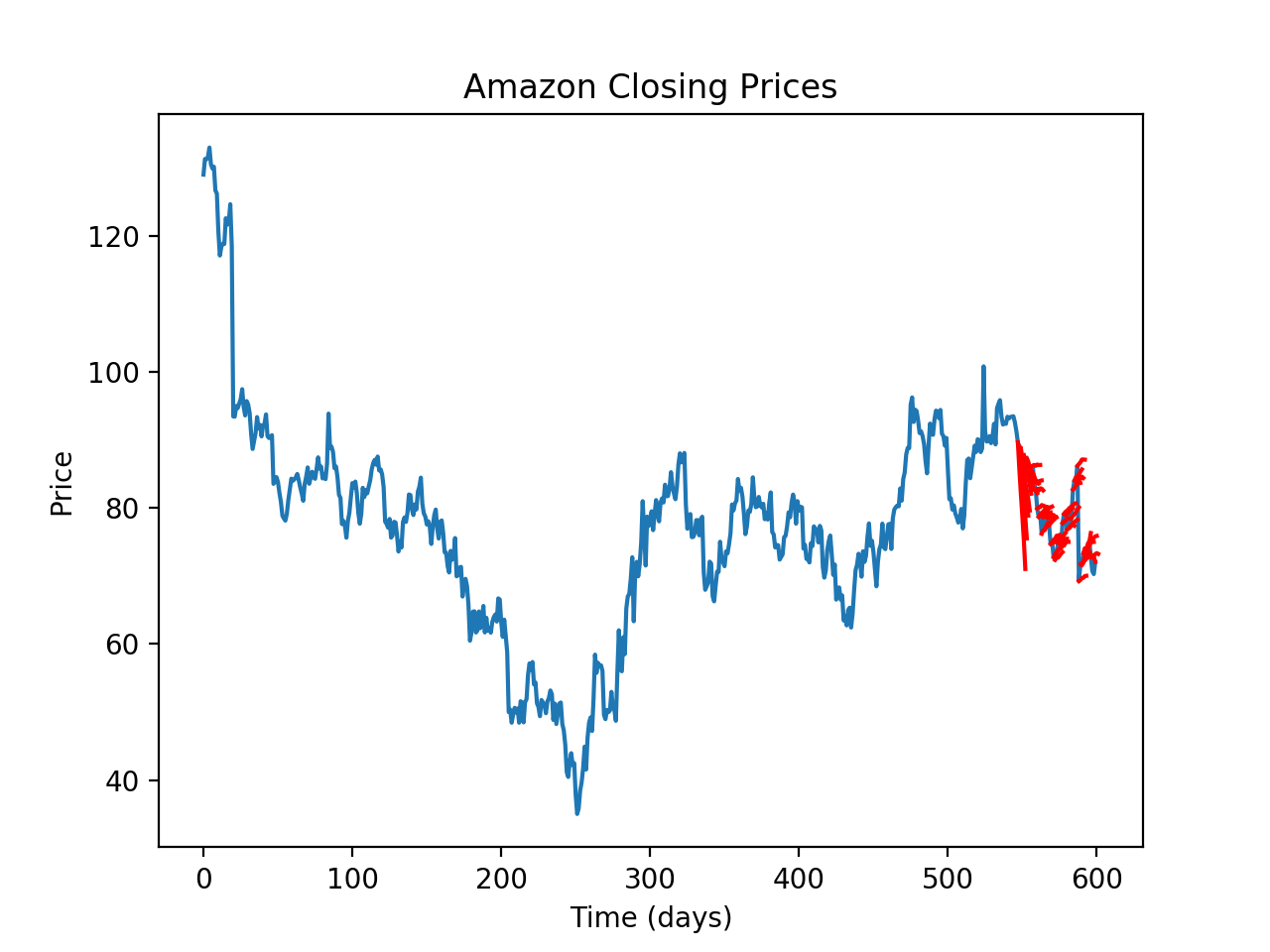}
\end{figure}
\end{center}

For a clearer look at the behavior of our predictions, we zoom in on the above graph and present a close-up view of our predictions on the test set shown below:

\begin{center}
\begin{figure}[h]
\includegraphics[width = 0.5\textwidth]{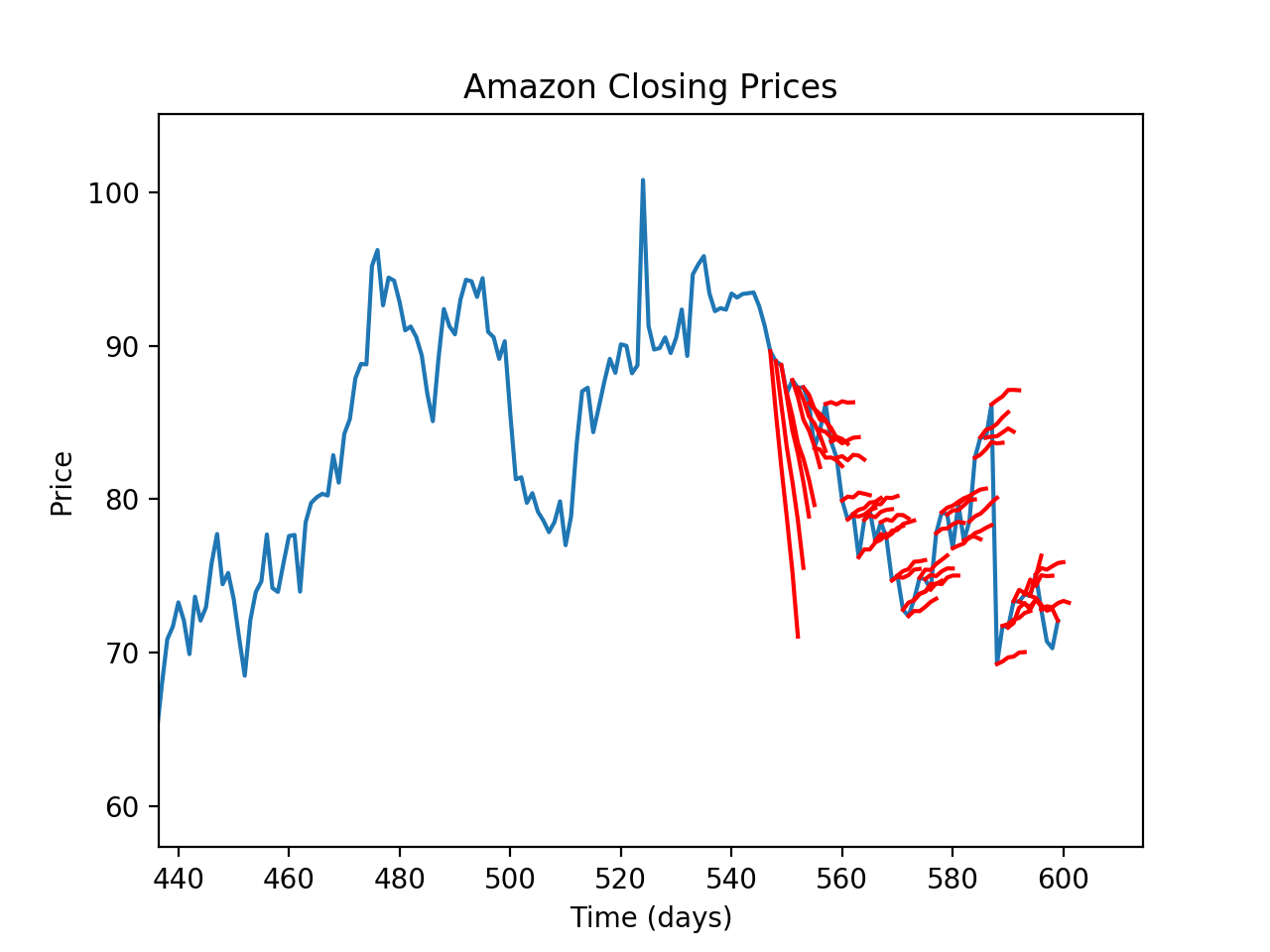}
\end{figure}
\end{center}

Here, we see an upward bias of our predictions - likely due to the fact that the AMZN saw a huge price increase in the previous 100 days leading up to this snapshot.

\subsection{Convolutional Network}
We did not find the same results as \cite{DLstock} from replication. The hyperparameters (especially the size of the hidden layers in the convolutions and the fully-connected) were not given in the work, which made optimizing them difficult.  

Roughly speaking, the parameters were: 
\begin{table}[h]
\caption{Hyper Parameters}
\label{parameters}
\begin{center}
\begin{tabular}{|c||c|}
\hline
Word-Embedding Vector Length & 100\\
\hline
Final Hidden Layer & 200\\
\hline
Short-Term Hidden Layer & 150 \\
\hline 
(Monthly, Weekly) Convolutional Hidden Layer & (40,20)\\
 \hline
 (Monthly, Weekly) Convolution Window Size & (3,3)\\
 \hline
 
 Optimizer & ADAM \\
 \hline
 (Day, Week, Month) Lags & (1,7,30)\\
 \hline
 (Batch Size and Learning Rate) & (50,0.001)\\
 \hline
 
\end{tabular}
\end{center}
\end{table}

\newpage 

The results are shown below when epoch size was 50:

\begin{figure}[h]
      \centering
     
      \includegraphics[scale=0.267]{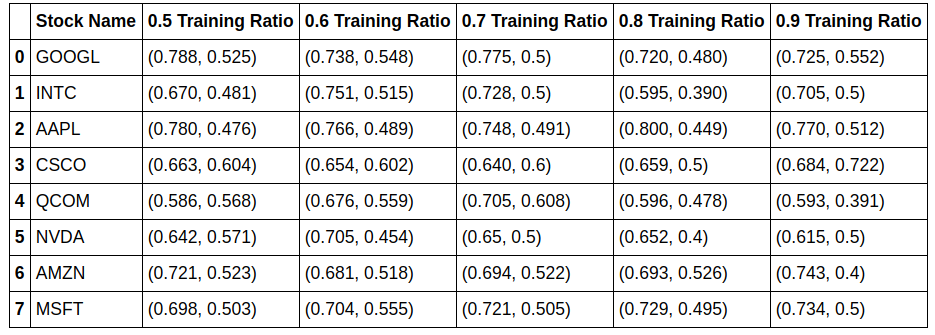}
      \caption{(Training Accuracy, Test Accuracy) for Full (Month, Week, Day) Network with epoch of 50.}
      \label{full_network}
      
      \includegraphics[scale=0.27]{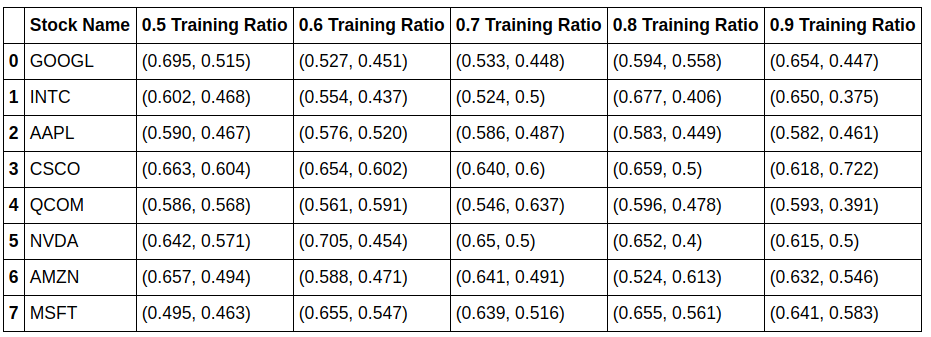}
      \caption{(Training Accuracy, Test Accuracy) for Short-Term (Day) Network Only with epoch of 50. }
      \label{short_network}
   \end{figure}

The epoch size was then increased to 500 to improve training accuracy:

\begin{figure}[h]
      \centering
     
      \includegraphics[scale=0.267]{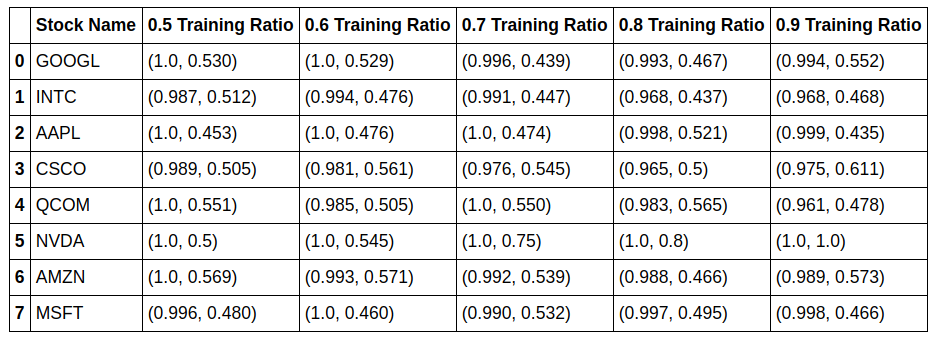}
      \caption{(Training Accuracy, Test Accuracy) for Full (Month, Week, Day) Network with shuffling with epoch of 500. }
      \label{full_noshuffle_500}
      
      \includegraphics[scale=0.27]{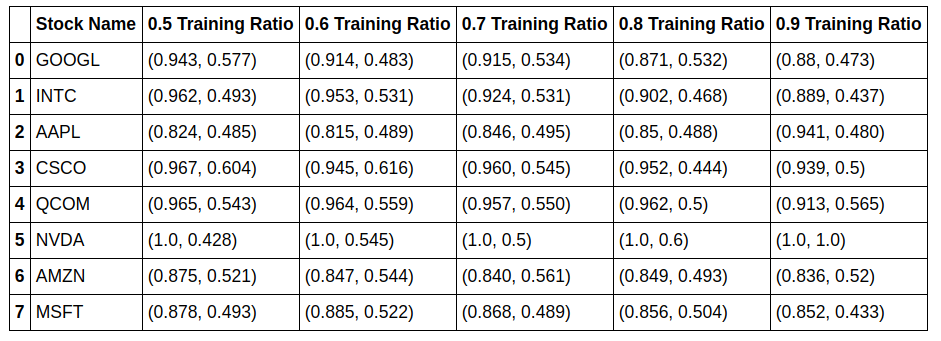}
      \caption{(Training Accuracy, Test Accuracy) for Short-Term (Day) Network Only with shuffling with epoch of 500. }
      \label{short_noshuffle_500}
   \end{figure}
  
From above, both networks dramatically improved their training accuracy. However, only the short-term network showed a consistent increase in testing accuracy. 
\newpage 
For the full-network, it did not necessarily improve testing accuracy, showing possibility of overfitting. Using a training ratio of $0.7$, we provided a graph (under no shuffling) for the relationship using the AAPL stock: 

\begin{figure}[h]
      \centering
     
      \includegraphics[scale=0.4]{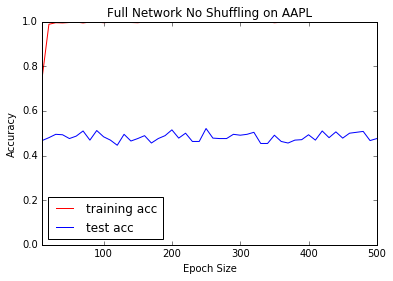}
      \caption{(Training Accuracy, Test Accuracy) for Full (Month, Week, Day) Network with no shuffling with varying number of epochs. }
      \label{full_noshuffle_500_graph}
      
      \includegraphics[scale=0.4]{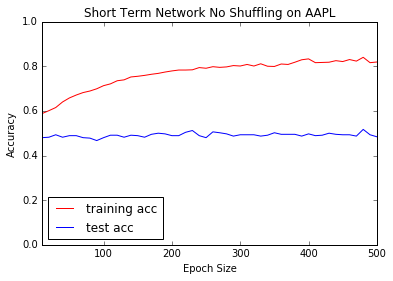}
      \caption{(Training Accuracy, Test Accuracy) for Short-Term (Day) Network Only with no shuffling varying number of epochs. }
      \label{short_noshuffle_500_graph}
   \end{figure}
  
It is clear that the full network can easily achieve close to 1.0 training accuracy very quickly and can strongly overfit. From the hyperparameters, the final hidden layer size asymptotically improved the training accuracy until there was a limit. Part of the reason for the full-network's overfitting could be due to lack of enough data points. We tested whether shuffling the training set made a difference as well, although experiments were first meant to be run sequentially. Furthermore, the LSTM-based models inherently uses the articles in chronological order. 

The result was inconclusive, as there was not a general trend of improvement. 
\newpage 
The results for epoch of 50 are shown:
\begin{figure}[h]
      \centering
     
      \includegraphics[scale=0.267]{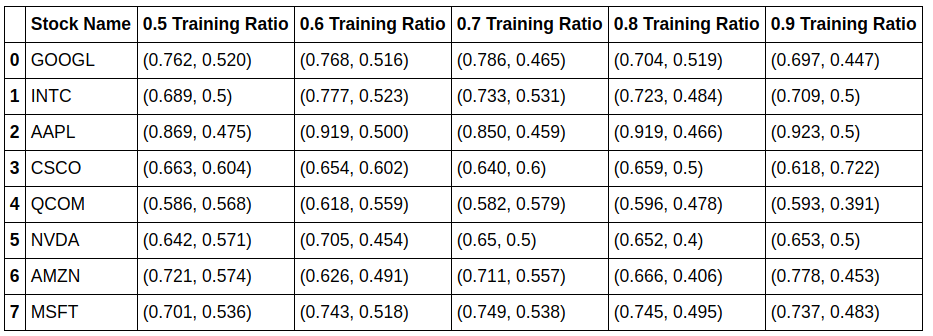}
      \caption{(Training Accuracy, Test Accuracy) for Full (Month, Week, Day) Network with shuffling with epoch of 50. }
      \label{full_shuffle_50}
      
      \includegraphics[scale=0.27]{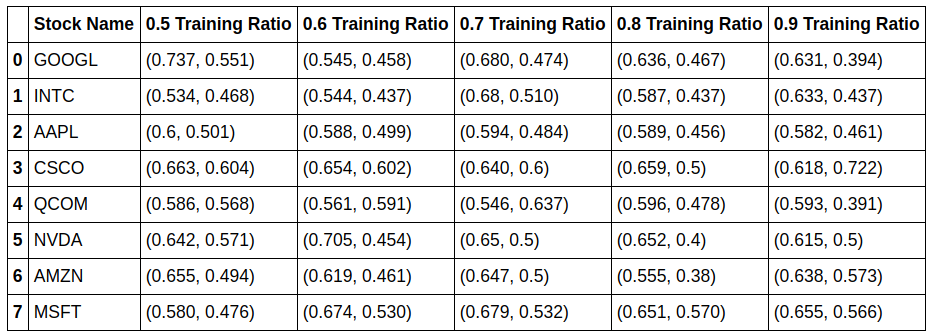}
      \caption{(Training Accuracy, Test Accuracy) for Short-Term (Day) Network Only with shuffling with epoch of 50. }
      \label{short_shuffle_50}
   \end{figure}

The results for an epoch of 500 are also shown:

\begin{figure}[h]
      \centering
     
      \includegraphics[scale=0.267]{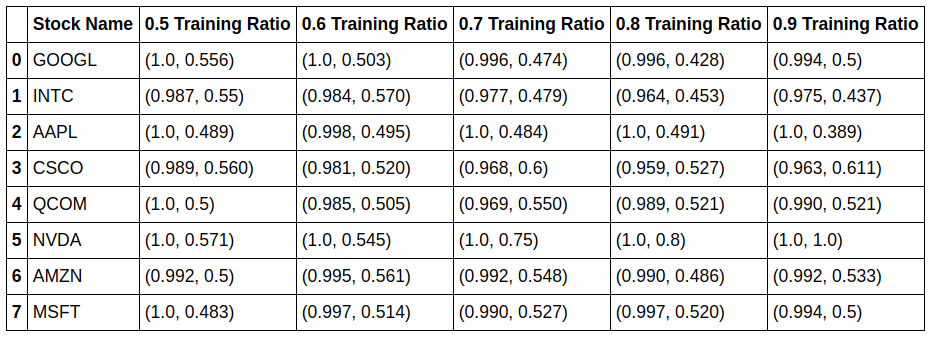}
      \caption{(Training Accuracy, Test Accuracy) for Full (Month, Week, Day) Network with shuffling with epoch of 500. }
      \label{full_shuffle_500}
      
      \includegraphics[scale=0.27]{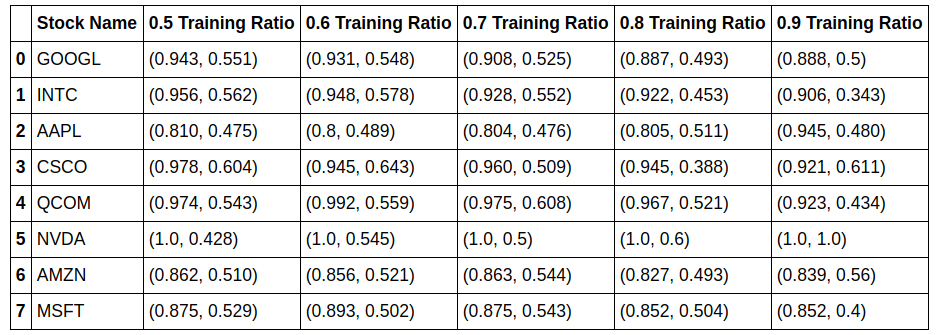}
      \caption{(Training Accuracy, Test Accuracy) for Short-Term (Day) Network Only with shuffling with epoch of 500. }
      \label{short_shuffle_500}
   \end{figure}

\newpage 

\subsection{LSTM-Based Event Embeddings}

We show learning curves (loss and accuracy) for our LSTM binary classifier on both the training and validation sets below:

\begin{center}
\begin{figure}[h]
\includegraphics[width = 0.5\textwidth]{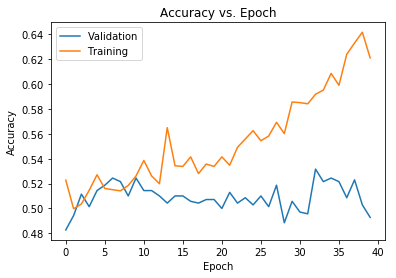}
\end{figure}
\end{center}

\begin{center}
\begin{figure}[h]
\includegraphics[width = 0.5\textwidth]{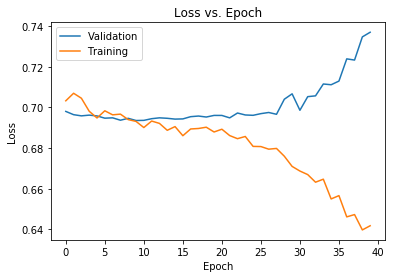}
\end{figure}
\end{center}
This shows that the LSTM is very prone to overfitting similar to the convolutional network - as epoch increases, so does training accuracy, but validation accuracy stays constant, or validation loss increases.

\section{Conclusions}
We summarized the available findings on the subject matter of individual stock prediction.

From our experiments, we noted essentially two patterns: 
\begin{itemize}
\item Even with the current batch of event data, it was very easy to overfit for both LSTM and convolutional network models, implying much more data is needed.
\item We did not achieve the strong results found in \cite{DLstock} with our convolutional re-implementations, which suggests either hyperparameters, data, or setup were slightly different. 
\end{itemize}
Overall therefore, it is unknown whether deep learning within the sentiment analysis context may help with individual stock prediction. Possible future approaches include:
\begin{itemize}
\item We hope to see future work given more news data, more varying stocks, and different architectures. The current experimentation setup overall produced prediction rates close to $50 \%$. 
\item It is possible that sentiment analysis on individual stocks may have better prediction rates with aspects of the stock other than price. For example, volume or liquidity may have better predictability because news articles are correlated with public opinion and tendency to trade. 
\end{itemize}
\addtolength{\textheight}{-12cm}   



\section{APPENDIX}
The code and data may be found in \url{https://github.com/Srizzle/Deep-Time-Series}. 
Essentially, the files are: 
\begin{itemize}
\item NN\_event\_to\_prediction.ipynb - This is the code for the neural network convolutional filter for up-down classification, found in \ref{conv}.
\item Stock\_Data\_Collection.ipynb - This collected daily stock prices and meta-data from Google Finance. 
\item nltk\_scores.ipynb - This uses NLTK's VADER package to produce mood-based sentiment scores of sentences and visualizations based on the data.
\item open\_ie.ipynb - This uses Stanford NLP's OpenIE extractor for event extraction on a sentence. 
\item event\_embedding.py - This produces event embeddings from the event tuple.
\item cnn\_classifier.py - Another version of the convolutional network from \cite{DLstock} in TensorFlow.
\item multistep\_pred.ipynb - Used VADER sentiment and previous prices for an LSTM.
\item sentence\_embedding\_prediction.ipynb - Used the event embeddings into an LSTM. 
\item \url{https://github.com/Srizzle/Deep-Time-Series/Papers} - Relevant Papers used in references. 
\end{itemize}
\section*{ACKNOWLEDGMENT}
We thank Francois Belletti for suggesting various links and resources, and Wilson Cai for suggesting the project.


\newpage 
\nocite{*}
\bibliographystyle{alpha} 
\bibliography{bib1}

\begin{thebibliography}{AYMU16}

\bibitem[AYMU16]{text}
Ryo Akita, Akira Yoshihara, Takashi Matsubara, and Kuniaki Uehara.
\newblock Deep learning for stock prediction using numerical and textual
  information, 06 2016.

\bibitem[Aza09]{sentimentthesis}
Pablo Azar.
\newblock Sentiment analysis in financial news, harvard bachelor's thesis,
  April 2009.

\bibitem[BMZ11]{twitter}
Johan Bollen, Huina Mao, and Xiao{-}Jun Zeng.
\newblock Twitter mood predicts the stock market.
\newblock {\em J. Comput. Science}, 2(1):1--8, 2011.

\bibitem[DL16]{DLnews}
Min-Yuh Day and Chia-Chou Lee.
\newblock Deep learning for financial sentiment analysis on finance news
  providers, 08 2016.

\bibitem[DZLD14]{structured}
Xiao Ding, Yue Zhang, Ting Liu, and Junwen Duan.
\newblock Using structured events to predict stock price movement: An empirical
  investigation.
\newblock In {\em Proceedings of the 2014 Conference on Empirical Methods in
  Natural Language Processing (EMNLP)}, pages 1415--1425. Association for
  Computational Linguistics, 2014.

\bibitem[DZLD15]{DLstock}
Xiao Ding, Yue Zhang, Ting Liu, and Junwen Duan.
\newblock Deep learning for event-driven stock prediction.
\newblock In {\em Proceedings of the 24th International Conference on
  Artificial Intelligence}, IJCAI'15, pages 2327--2333. AAAI Press, 2015.

\bibitem[HG14]{VADER}
Clayton~J. Hutto and Eric Gilbert.
\newblock {VADER:} {A} parsimonious rule-based model for sentiment analysis of
  social media text.
\newblock In {\em Proceedings of the Eighth International Conference on Weblogs
  and Social Media, {ICWSM} 2014, Ann Arbor, Michigan, USA, June 1-4, 2014.},
  2014.

\bibitem[HPW16]{deep-learning-finance}
J.~B. Heaton, N.~G. Polson, and J.~H. Witte.
\newblock Deep learning in finance, 02 2016.

\bibitem[IN17]{limited}
Vasileios Iosifidis and Eirini Ntoutsi.
\newblock Large scale sentiment learning with limited labels.
\newblock In {\em Proceedings of the 23rd ACM SIGKDD International Conference
  on Knowledge Discovery and Data Mining}, KDD '17, pages 1823--1832, New York,
  NY, USA, 2017. ACM.

\bibitem[PCT08]{words}
Sofus~Macskassy Paul C.~Tetlock, Maytal Saar-Tsechansy.
\newblock More than words: Quantifying language to measure firms' fundamentals.
\newblock In {\em The Journal of Finance}, 2008.

\bibitem[RACM15]{twitterPLOS}
Gabriele Ranco, Darko Aleksovski, Guido Caldarelli, and Igor Mozetic.
\newblock Investigating the relations between twitter sentiment and stock
  prices.
\newblock {\em CoRR}, abs/1506.02431, 2015.

\bibitem[RDM12]{news_short}
Kira Radinsky, Sagie Davidovich, and Shaul Markovitch.
\newblock Learning causality for news events prediction.
\newblock In {\em Proceedings of the 21st International Conference on World
  Wide Web}, WWW '12, pages 909--918, New York, NY, USA, 2012. ACM.

\bibitem[RPS09]{textual-analysis}
Hsinchun~Chen Robert P.~Schumaker.
\newblock Textual analysis of stock market prediction using breaking financial
  news.
\newblock In {\em ACM Transactions on Information Systems}, 02 2009.

\bibitem[RS12]{sentiment}
Tushar Rao and Saket Srivastava.
\newblock Analyzing stock market movements using twitter sentiment analysis.
\newblock In {\em Proceedings of the 2012 International Conference on Advances
  in Social Networks Analysis and Mining (ASONAM 2012)}, ASONAM '12, pages
  119--123, Washington, DC, USA, 2012. IEEE Computer Society.

\bibitem[Tet07]{media}
Paul~C. Tetlock.
\newblock Giving content to investor sentiment: The role of media in the stock
  market.
\newblock In {\em The Journal of Finance}, 2007.

\end{thebibliography}

\end{document}